\documentclass{article}

\PassOptionsToPackage{numbers,compress}{natbib}

 \usepackage[preprint]{neurips_2026}

\usepackage[utf8]{inputenc}
\usepackage[T1]{fontenc}
\usepackage{hyperref}
\usepackage{url}
\usepackage{booktabs}
\usepackage{amsfonts}
\usepackage{nicefrac}
\usepackage{microtype}
\usepackage{xcolor}
\usepackage{amsmath}           
\usepackage{amssymb}           
\usepackage{bm}                
\usepackage{graphicx}          
\usepackage{algorithm}         
\usepackage{algpseudocode}     
\usepackage[table]{xcolor}
\usepackage{makecell}
\usepackage{bbding}
\usepackage{multirow}
\usepackage{soul}
\usepackage{algorithm}
\usepackage{algorithmicx}
\usepackage{algpseudocode}
\usepackage{tikz}
\usepackage{pgfplots}
\pgfplotsset{compat=1.18}
\usepgfplotslibrary{groupplots}
\newcommand{\GSDiff}{\textsc{GS-Diff}}

\title{From Pixels to Primitives: Scene Change Detection in 3D Gaussian Splatting}

\author{%
  Chamuditha Jayanga Galappaththige$^{1,2}$ \hspace{10pt} Jason Lai$^{3}$ \hspace{10pt}  Timothy Patten$^{2,4}$ \\ \textbf{Donald Dansereau}$^{2,3}$  \hspace{10pt}   \textbf{Niko Suenderhauf}$^{1,2}$ \hspace{10pt}  \textbf{Dimity Miller}$^{1,2}$ \\
  $^1$QUT Centre for Robotics \hspace{1pt} $^2$ARIAM \hspace{1pt} $^3$ACFR, University of Sydney \hspace{1pt} $^4${Abyss Solutions}\\
  \texttt{\{chamuditha.galappaththige, d24.miller\}@qut.edu.au} \\
}
\begin{document}

\maketitle

\begin{abstract}
 \label{sec:abs}
Scene change detection methods built on Gaussian splatting universally follow a render-then-compare paradigm: the pre-change scene is rendered into 2D and compared against post-change images via pixel or feature residuals. This \emph{change detection problem with Gaussian Splatting has been treated as a question about pixels; we treat it as a question about primitives.} We provide direct evidence that native primitive attributes alone -- position, anisotropic covariance, and color -- carry sufficient signal for scene change detection. What makes primitive-space comparison hard is the under-constrained nature of Gaussian splatting representation: independent optimizations yield primitive solutions whose count, positions, shapes, and colors differ even where nothing has changed. We address this challenge with anisotropic models of geometric and photometric drift, complemented by a per-primitive observability term that reflects the extent to which each Gaussian is constrained by the camera geometry. Operating directly on primitives gives our method, \GSDiff{}, two properties that distinguish it from render-then-compare methods. First, change maps are multi-view consistent by construction, where prior work had to learn this through an additional optimization objective. Second, geometric and appearance changes are scored separately, identifying not just \emph{where} but \emph{what kind of} change occurred, distinguishing structural changes (e.g., an added object) from surface-level ones (e.g., a color change) without supervision or external model dependencies. On real-world benchmarks, \GSDiff{} surpasses the prior state-of-the-art approach by $\sim\textbf{17\%}$ in mean Intersection over Union. Code and annotations will be released upon acceptance.
 
\end{abstract}

\section{Introduction}
\label{sec:Intro}

3D Gaussian splatting (3DGS)~\cite{kerbl20233d} has become a foundational representation for photorealistic digital twins~\cite{gao2025digital}, with adoption spanning spatial maps~\cite{matsuki2024gaussian}, heritage preservation~\cite{dahaghin2024gaussian}, and industrial asset management~\cite{galappaththige2025multi}. A critical challenge in these applications is understanding how environments evolve over time: an agent visits a scene at different times, capturing images from unconstrained viewpoints, and must detect and localize changes that have occurred between inspections. This \emph{Multi-View Scene Change Detection} (SCD) problem~\cite{galappaththige2025changes,galappaththige2025multi,lu20253dgs} requires robustness to arbitrary camera trajectories and the ability to distinguish genuine structural or appearance changes from irrelevant variations such as lighting shifts, shadows, and reflections.

\begin{figure}[t]
  \centering
  \includegraphics[width=\linewidth]{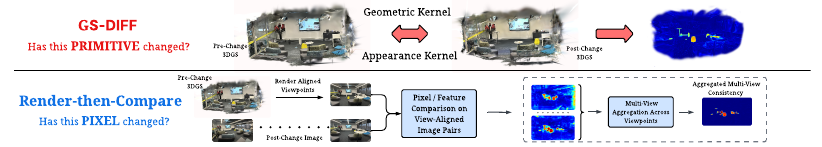}
  \vspace{-12pt}
\caption{From pixels to primitives. Prior multi-view SCD methods question \emph{pixels}, comparing rendered viewpoint pairs in image space and learning to aggregate change evidence scattered across viewpoints (bottom). \GSDiff{} questions \emph{primitives}, comparing two 3DGS reconstructions directly in primitive space (top). Multi-view consistency emerges \emph{by construction} from the shared 3D representation, eliminating both per-view comparison and learned aggregation.}
  \label{fig:paradigm}
  \vspace{-8pt}
\end{figure}

Existing work~\cite{galappaththige2025multi, lu20253dgs, jiang2025gaussian, kruse2024splatpose, liu2024splatpose+, ackermann2025clsplats, galappaththige2025changes} follows a \emph{render-then-compare} paradigm, treating 3DGS as a \emph{rendering engine}: the pre-change scene is reconstructed, rendered at post-change poses, and compared to post-change images via foundation-model features~\cite{oquab2023dinov2, ravi2024sam}, pixel-level metrics~\cite{wang_image_2004}, or combinations thereof (Figure~\ref{fig:paradigm}). Because genuine changes register consistently across views while distractors such as shadows and reflections fire inconsistently, state-of-the-art methods aggregate change cues across views to enforce multi-view consistency~\cite{galappaththige2025multi, galappaththige2025changes, lu20253dgs, zhou2026d, ackermann2025clsplats}, suppressing these inconsistencies. In every case, the \emph{change signal} originates in \emph{image space}; primitives serve only as a rendering and aggregation substrate, never as the basis for comparison.

Prior work compares scenes in image space -- pixel by pixel, view by view. We instead ask whether the comparison can happen directly between Gaussian primitives.
Each primitive explicitly encodes position, covariance, opacity, and color -- the very attributes that change when a scene changes, and determine how those changes appear in a rendered image. If scene-level change is already encoded in the primitives, we hypothesize that change can be detected in the 3DGS representation directly, removing both the detour through rendered images and the cross-view aggregation it requires.

To this end, we present \GSDiff{}, which detects scene-level changes by comparing two independently reconstructed 3DGS scenes \emph{directly in primitive space}. What has -- so far -- prevented direct primitive-space comparison is the under-constrained nature of 3DGS reconstruction~\cite{kerbl20233d}: for a given set of images, the representation is non-unique and many distinct primitive configurations, differing in count, position, shape, and color, explain the same observations. Therefore, independent optimizations of even unchanged scenes converge to solutions without direct primitive-to-primitive correspondence. We call the resulting geometric and color differences between corresponding unchanged regions \emph{geometric} and \emph{photometric drift}. We address this drift in two ways to enable direct primitive comparison: a geometric kernel over position and covariance, complemented by a Fisher Information observability term based on camera view geometry, absorbs geometric drift (\S\ref{sec:geo}), and an appearance kernel over diffuse color absorbs photometric drift (\S\ref{sec:app}).

Operating directly in primitive space gives \GSDiff{} two appealing structural properties. 
First, the change signal is multi-view consistent \emph{by construction}, whereas prior work~\cite{galappaththige2025multi,galappaththige2025changes} learned viewpoint consistency through multi-view aggregation of 2D change cues. 
Second, by considering primitive attributes separately -- position-covariance versus color -- \GSDiff{} identifies not only \emph{where} change has occurred but \emph{what kind of} change it is. This disambiguation is unavailable to previous image-space methods without learned change classification or auxiliary models.
To quantitatively evaluate this, we annotate each change pixel in the PASLCD benchmark~\cite{galappaththige2025multi} as structural or surface-level.

\noindent In summary, we make the following claims, validated by our experiments:
\begin{itemize}

 \item \textbf{Gaussian primitives alone carry sufficient signal for state-of-the-art multi-view SCD.} We present \GSDiff{}, the first method to perform SCD in 3DGS primitive space, surpassing the strongest image-space baseline by $\sim17\%$ in mIoU (see \S\ref{sec:main_results}).

    \item \textbf{Reconstruction drift is the obstacle to direct primitive comparison.} We resolve it with anisotropic geometric and data-driven photometric drift models enabling robust comparison across independent 3DGS reconstructions (see \S\ref{sec:ablations}).
    
     \item\textbf{Primitive-space comparison allows distinguishing structural from surface-only changes.} This is inherent to \GSDiff{} and requires no supervision or auxiliary models (see \S\ref{sec:disambig}).
\end{itemize}

\section{Related Work}
\label{sec:RelatedWork}

\paragraph{Scene Change Detection via Image Comparison.} SCD across unconstrained viewpoints has progressed through three eras, all operating in image space and depending on learned or pretrained image features. Early methods assume strictly paired pre- and post-change views~\cite{alcantarilla2018street, daudt2018fully, varghese2018changenet, sakurada_change_2015}. A second line tolerated minor viewpoint variation through supervised annotations~\cite{cyws2d, cyws3d}, but supervised formulations are inherently brittle: dense annotation is expensive, the distribution of real-world changes is unbounded, and performance degrades~\cite{kim2025towards, cho2025zero} sharply under distribution shift~\cite{galappaththige2024towards,gulrajani2020search}. The most recent line accordingly pivots to label-free and zero-shot paradigms~\cite{lin2025robust,kim2025towards, cho2025zero, kannanZero, alpherts2025emplace, friedlander2026goldilocs} driven by vision foundation models~\cite{oquab2023dinov2, ravi2024sam}. Across all three eras, the comparison remains structurally tethered to 2D image space and assumes viewpoint consistency between pairs -- a condition rarely satisfied when observing agents follow independent trajectories.

Recent methods enable unconstrained viewpoints by leveraging 3DGS~\cite{kerbl20233d} to synthesize reference views at inference-time poses. Object-level pose-agnostic anomaly detection~\cite{kruse2024splatpose, liu2024splatpose+} scores changes by rendering views and then comparing features, while the SCD community extended this to complex multi-change scenes~\cite{lu20253dgs, jiang2025gaussian, galappaththige2025multi,zhou2026d,galappaththige2026predictive}. A core insight in this line is that per-view change signals are not consistent across viewpoints: true changes register with varying strength depending on visibility, lighting, and feature response, but consistently across viewpoints, while distractors (e.g., shadows or reflections) fire inconsistently. MV3DCD~\cite{galappaththige2025multi} embedded change cues as per-Gaussian change channels and fused viewpoints through hard thresholding and heuristics, which enabled MV3DCD to suppress view-inconsistent signals that pairwise methods could not. O-SCD~\cite{galappaththige2025changes} replaced this heuristic fusion with a self-supervised loss, providing evidence that the strength of the learned objective is itself a ceiling on achievable consistency.
However, all existing methods use Gaussian primitives as a mere substrate for 2D rendering or aggregation, ultimately deriving 2D change signals in image space using pretrained features, a process susceptible to the chosen backbone's failure modes~\cite{galappaththige2025multi}.
We break from this paradigm by performing direct primitive-space comparison. This shift, enabled by explicit reconstruction drift modeling, yields a pipeline that is free of foundation model dependence, change annotations, and extended multi-view aggregation machinery.

\paragraph{Scene Change Detection on 3D Representations.}
Change detection between 3D representations has a long history in geospatial surveying, comparing bitemporal LiDAR scans via cloud-to-cloud distances~\cite{c2c}, M3C2~\cite{LAGUE201310}, octree occupancy~\cite{gehrung2019fast}, scene graphs~\cite{looper20233d}, and deep learning on point clouds~\cite{yew2021city, de2023siamese, qiu2023pointcloud}. Our method also compares two 3D representations directly, but differs in two critical respects. First, point cloud methods operate on raw LiDAR data with sensor-characterized noise, whereas our representation is optimized from images and introduces reconstruction drift with no LiDAR analogue. Second, our method exploits the anisotropic spatial extent of Gaussian primitives, which point clouds do not carry. Images encode both geometry and photometric appearance, and 3DGS inherits this richness, admitting joint geometric-photometric scoring over a shared representation. As 3DGS becomes the standard format for repeated spatial capture~\cite{gao2025digital}, native change detection on Gaussian primitives is timely; ours is the first such approach.

\section{Method}
\label{sec:Method}

\subsection{Preliminaries}
\label{sec:prelim}

\paragraph{Task setting.} 
Following prior work~\cite{lu20253dgs,zhou2026d,galappaththige2025multi, galappaththige2025changes}, a scene is captured at two different times, producing a \emph{reference} (pre-change) image set $\mathcal{I}^{(1)}$ and an \emph{inference} (post-change) image set $\mathcal{I}^{(2)}$ along independent camera trajectories that need not share identical views. The objective is to output a binary change mask for images in $\mathcal{I}^{(2)}$, marking pixels whose scene content differs from $\mathcal{I}^{(1)}$.

\paragraph{Gaussian Splatting.} 3DGS~\cite{kerbl20233d} represents a scene as a set of anisotropic Gaussians $\mathcal{G} = \{g_i\}_{i=1}^{N}$ optimized from posed images. Each primitive $g_i$ is parameterized by a position $\bm{\mu}_i \in \mathbb{R}^{3}$, a covariance $\bm{\Sigma}_i = \mathbf{R}_i \mathbf{S}_i \mathbf{S}_i^{\top} \mathbf{R}_i^{\top}$ with rotation $\mathbf{R}_i \in SO(3)$ and scales $\mathbf{S}_i = \mathrm{diag}(s_1, s_2, s_3)$, an opacity $o_i$, and a spherical-harmonic (SH) color whose DC coefficient $\mathbf{c}_i \in \mathbb{R}^{3}$ encodes diffuse color while higher-order coefficients capture view-dependent effects. Images are rendered by splatting each primitive to the image plane and alpha-compositing~\cite{zwicker2002ewa}. For geometrically accurate reconstructions, the surface normal $\mathbf{n}_i \in \mathbb{S}^2$ is approximated by the column of $\mathbf{R}_i$ corresponding to the smallest axis of $\mathbf{S}_i$.

\subsection{\GSDiff{}: Primitive-Space Scene Change Detection in 3D Gaussian Splatting}
\label{sec:method-core}
\begin{figure}[t]
  \centering
  \includegraphics[width=\linewidth]{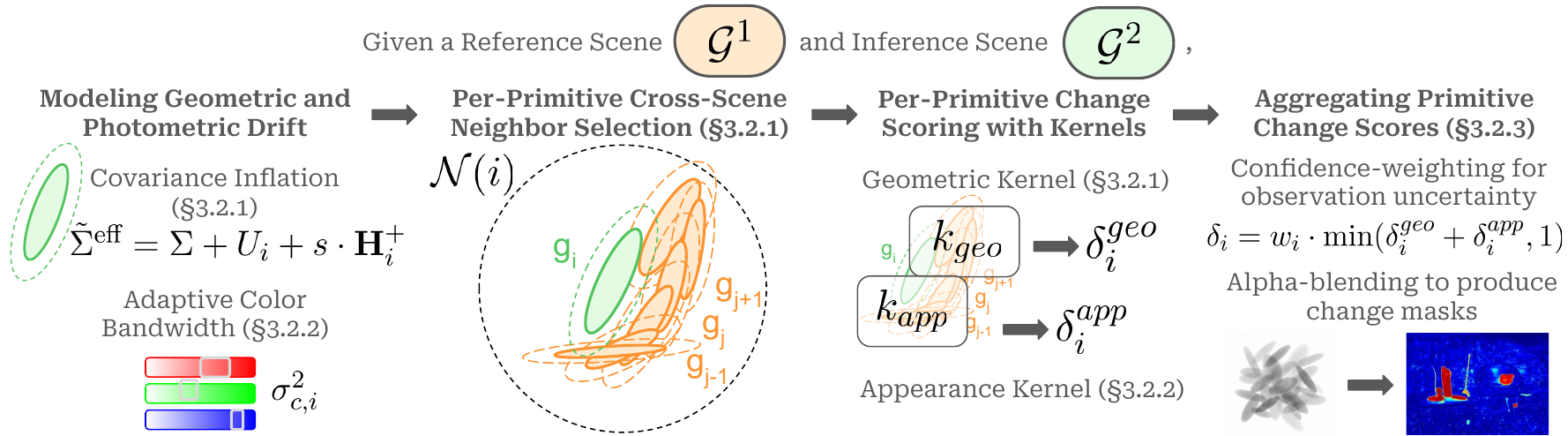}
  \vspace{-12pt}
  \caption{The \GSDiff{} pipeline: We model the expected geometric and photometric drift between 3DGS representations, using the inflated covariance of each primitive to find its cross-scene neighbor set. A geometric kernel and appearance kernel evaluate change over the neighbor set to compute drift-aware change scores. These change scores are combined and weighted by observation uncertainty, and can then be rendered as change score maps for any viewpoint.}
  \label{fig:pipeline}
\end{figure}

Figure~\ref{fig:pipeline} illustrates the core idea of \GSDiff{}. Given two reconstructed scenes, the central challenge to direct comparison is non-uniqueness in the 3DGS representation: independent reconstructions of scenes yield many plausible primitive solutions differing in position, shape, and color while describing the same observations even when unchanged. \GSDiff{} first builds an anisotropic covariance inflation matrix for each primitive, capturing geometric drift that arises from both representation ambiguity and observation uncertainty, and folds it into each primitive's own covariance matrix. With this effective covariance accommodating geometric drift, a plausible cross-reconstruction neighbor set is drawn. Then a geometric kernel evaluates how well the existence of that primitive is described within the neighbor set, accommodating the geometric drift (\S\ref{sec:geo}). An appearance kernel then operates on the same neighbor set, evaluating how well the diffuse color is described (\S\ref{sec:app}). To produce a change mask for any queried viewpoint, each primitive score is weighted by a confidence term and rendered following alpha-compositing~\cite{kerbl20233d}. We can further consider scores separately to disambiguate differences arising from underlying structural versus surface-level changes (\S\ref{sec:agg}).

\paragraph{Reconstructing 3DGS Representations.} Following prior multi-view SCD work~\cite{galappaththige2025multi,lu20253dgs,zhou2026d,galappaththige2025changes}, we use COLMAP~\cite{schonberger2016structure} to estimate camera poses $\mathcal{C}^{(1)}$ and $\mathcal{C}^{(2)}$ for the image sets $\mathcal{I}^{(1)}, \mathcal{I}^{(2)}$ in a common coordinate frame. From $(\mathcal{I}^{(1)}, \mathcal{C}^{(1)})$ and $(\mathcal{I}^{(2)}, \mathcal{C}^{(2)})$ we then independently build two geometrically accurate reconstructions~\cite{fastgs, chen2024pgsr} $\mathcal{G}^{(1)} = \{g_i^{(1)}\}_{i=1}^{N_1}$ and $\mathcal{G}^{(2)} = \{g_j^{(2)}\}_{j=1}^{N_2}$. Since change can only be assessed in regions both image sets observed, we discard primitives visible in only one of the camera frustum sets $\mathcal{C}^{(1)}, \mathcal{C}^{(2)}$ before comparison begins. We assume $\mathcal{I}^{(1)}, \mathcal{I}^{(2)}$ individually provide sufficient scene coverage for geometrically accurate reconstruction.

\subsubsection{Neighbor Retrieval and Geometric Scoring Under Geometric Drift}
\label{sec:geo}
The \emph{geometric kernel} scores the change in local geometry between reconstructions. A na\"ive nearest-neighbor (NN) distance based only on primitive position is brittle, as we empirically show in \S~\ref{sec:ablations} -- NN ignores the spatial extent of each primitive and is sensitive to geometric drift; the same observations can be described by different configurations of Gaussians. Our key insight is that the primitive's spatial extent, captured by its covariance, can be leveraged to absorb this drift. We attribute drift to two sources and model each as an additive, anisotropic covariance inflation. The resulting effective covariance is used both to retrieve cross-reconstruction neighbors and regularize the kernel.

\paragraph{Sources of geometric drift.} The first source is representation ambiguity. 3DGS representation is non-unique for a given scene: many primitive configurations differing in count, position, and shape render to the same observed images, and independent optimizations converge to different primitive solutions even on unchanged regions. The resulting positional differences are anisotropic: with sufficient camera baseline, multi-view triangulation tightly constrains across-surface (normal) position and ambiguity dominates tangentially, while under-constrained baselines admit larger normal-direction drift. The decomposition into $u_n$ and $u_t$ in Eq.~\ref{eq:repamb} captures whichever asymmetry holds. 

The second source is observation uncertainty, a form of epistemic uncertainty~\cite{kendall2017uncertainties,hullermeier2021aleatoric} that grows when a primitive is poorly constrained by the camera viewing geometry. A primitive observed from a single narrow baseline, for example, has its position only weakly determined along the viewing ray, and small differences in optimization initialization could shift it substantially. Without separating poorly observed primitives from genuinely changed ones, both register as differences.

\paragraph{Modeling representation ambiguity.}
Representation ambiguity is a property of the reconstruction process itself; its scale can be estimated from the data. For each primitive $g_i^{(1)} \in \mathcal{G}^{(1)}$ we find its Euclidean nearest neighbor $g_{j^*}^{(2)} \in \mathcal{G}^{(2)}$ and decompose the displacement $\bm{\Delta}_i = \bm{\mu}_{j^*}^{(2)} - \bm{\mu}_i^{(1)}$ into normal and tangential components relative to the surface normal $\mathbf{n}_i$:
\begin{equation}
d_n^{(i)} = \left| \mathbf{n}_i^{\top} \bm{\Delta}_i \right|, \qquad
d_t^{(i)} = \left\| \bm{\Delta}_i - (\mathbf{n}_i^{\top} \bm{\Delta}_i)\mathbf{n}_i \right\|_2.
\end{equation}
We summarize the per-direction drift scale by upper-quartile statistics $Q_{0.75}$, which avoids contamination from the tail of the distribution that contains genuinely changed regions:
\begin{equation}
u_n^2 = \left[Q_{0.75}\!\left(\{d_n^{(k)}\}_{k \in |\mathcal{G}|}\right)\right]^2, \qquad
u_t^2 = \left[Q_{0.75}\!\left(\{d_t^{(k)}\}_{k \in |\mathcal{G}|}\right)\right]^2, 
\end{equation}

computed bidirectionally ($\mathcal{G}^{(1)} \to \mathcal{G}^{(2)}$ and $\mathcal{G}^{(2)} \to \mathcal{G}^{(1)}$) and averaged for symmetry. We assemble an anisotropic inflation matrix and add it to each primitive's covariance:
\begin{equation}
\mathbf{U}_i = u_t^2 \, \mathbf{I}_3 + (u_n^2 - u_t^2) \, \mathbf{n}_i \mathbf{n}_i^{\top},
\qquad
\tilde{\bm{\Sigma}}_i = \bm{\Sigma}_i + \mathbf{U}_i.
\label{eq:repamb}
\end{equation}
Effective covariance $\tilde{\bm{\Sigma}}_i$ now reflects both each primitive's local extent and the typical geometric drift.

\paragraph{Modeling observation uncertainty.}
Observation uncertainty is intrinsically per-primitive: a primitive seen from many angles is well-constrained, while one seen from a narrow baseline is not. We quantify it through a Fisher information matrix (FIM) 
accumulated over each primitive's frustum-visible cameras $\mathcal{C}_i^{\text{vis}}$, following the standard form for inverse-depth uncertainty in multi-view geometry~\cite{Hartley:2003:MVG:861369}:
\begin{equation}
\mathbf{H}_i = \sum_{c \in \mathcal{C}_i^{\text{vis}}} \frac{1}{\left\| \bm{\mu}_i - \mathbf{o}_c \right\|^2} \left( \mathbf{I}_3 - \mathbf{v}_{i,c} \mathbf{v}_{i,c}^{\top} \right),
\label{eq:fim}
\end{equation}
where $\mathbf{o}_c$ is the camera center, $d_{i,c} = \|\bm{\mu}_i - \mathbf{o}_c\|$ the distance from primitive $i$ to camera $c$, and $\mathbf{v}_{i,c} = (\bm{\mu}_i - \mathbf{o}_c)/d_{i,c}$ the unit viewing ray. The projection $\mathbf{I}_3 - \mathbf{v}\mathbf{v}^{\top}$ encodes that each camera constrains a primitive in the image plane but not along the viewing direction; the $1/d_{i,c}^2$ factor encodes perspective fall-off. A primitive observed from many angles at close range yields a well-conditioned $\mathbf{H}_i$; one observed from a single viewpoint yields a near-singular $\mathbf{H}_i$ with one unconstrained direction. We inject the pseudo-inverse $\mathbf{H}_i^{+}$ into the covariance with a data-driven scale:
\begin{equation}
\tilde{\bm{\Sigma}}_i^{\text{eff}} = \tilde{\bm{\Sigma}}_i + s \cdot \mathbf{H}^{+}_i,
\qquad
s = \frac{\mathrm{median}\!\left(\{\mathrm{tr}(\tilde{\bm{\Sigma}}_k)\}_{k \in |\mathcal{G}|}\right)}{\mathrm{median}\!\left(\{\mathrm{tr}(\mathbf{H}_k^{+})\}_{k \in |\mathcal{G}|}\right)}.
\label{eq:obsunc}
\end{equation}
The scale $s$ matches the typical magnitude of $\mathbf{H}_i^{+}$ to that of the representation-ambiguity inflation, so neither dominates. Where a primitive is under-constrained along a particular direction, $\mathbf{H}_i^{+}$ has a large eigenvalue along that direction, and $\tilde{\bm{\Sigma}}_i^{\text{eff}}$ stretches accordingly -- the kernel will tolerate larger displacements there before declaring a change. The same FIM also informs a per-primitive confidence weight at render time (\S\ref{sec:agg}), giving observation uncertainty two complementary roles.

\paragraph{Neighbor retrieval and geometric kernel.}
With $\tilde{\bm{\Sigma}}_i^{\text{eff}}$ encoding the total expected geometric drift for primitive $i$, we retrieve a cross-reconstruction neighbor set within an drift-aware search radius:
\begin{equation}
\mathcal{N}(i) = \left\{ j : \left\| \bm{\mu}_i^{(1)} - \bm{\mu}_j^{(2)} \right\| \leq \eta \cdot \sqrt{\lambda_{\max}\!\left(\tilde{\bm{\Sigma}}_i^{\text{eff}}\right)} \right\}, \quad \eta = 3.
\label{eq:neighbors}
\end{equation}
The Euclidean ball is a conservative superset of the Mahalanobis $\eta$-ellipsoid, ensuring no candidate within Mahalanobis distance $\eta$ is missed while permitting fast spatial indexing. To measure geometric similarity between a primitive and its cross-reconstruction neighbor set, we then apply the geometric kernel as an unnormalized anisotropic Mahalanobis radial basis function (RBF):
\begin{equation}
k_{\text{geo}}(g_i, g_j) = \exp\!\left( -\tfrac{1}{2} (\bm{\mu}_i - \bm{\mu}_j)^{\top} \mathbf{M}_{ij}^{-1} (\bm{\mu}_i - \bm{\mu}_j) \right),
\qquad \mathbf{M}_{ij} = \tilde{\bm{\Sigma}}_i^{\text{eff}} + \tilde{\bm{\Sigma}}_j^{\text{eff}}.
\label{eq:kgeo}
\end{equation}
We omit the normalization prefactor $(2\pi)^{-3/2}|\mathbf{M}_{ij}|^{-1/2}$: the determinant penalizes co-located primitives whose covariance magnitudes differ, a routine artifact of independent reconstructions with different densification histories. The unnormalized form preserves $k_{\text{geo}} = 1$ at $\bm{\mu}_i = \bm{\mu}_j$ regardless of covariance scale; we ablate this choice in \S\ref{sec:ablations}. The per-primitive geometric change score is:
\begin{equation}
\delta_i^{\text{geo}} = 1 - \max_{j \in \mathcal{N}(i)} k_{\text{geo}}(g_i, g_j).
\label{eq:dgeo}
\end{equation}
A high $\delta_i^{\text{geo}}$ indicates that no primitive in the neighborhood $\mathcal{N}(i)$ set explains the existence of $g_i$ once drift has been absorbed, which we interpret as evidence of geometric change.

\subsubsection{Appearance Scoring Under Photometric Drift}
\label{sec:app}

After capturing geometric change, we then apply the \emph{appearance kernel} to score color change between each primitive and its geometric neighbor set $\mathcal{N}(i)$ from \S\ref{sec:geo}. A naive direct comparison of DC colors is again brittle (as we empirically show in \S~\ref{sec:ablations}), this time due to \emph{photometric drift}: two reconstructions of an unchanged scene yield primitives whose DC colors differ even at matched positions. Two factors contribute. First, the spherical-harmonic decomposition of color is itself non-unique and subject to representation ambiguity; different SH configurations can produce the same diffuse appearance, so lighting and residual view-dependent effects bake into the DC term in inconsistent ways across reconstructions. Second, two captures rarely share identical illumination; ambient changes, time-of-day shifts, and automatic exposure or white-balance drift introduce a global color offset that is absorbed into the DC coefficients.

\paragraph{Modeling photometric drift.}
From the data, we estimate a color bandwidth $\sigma_c$ as the typical photometric drift magnitude. For each primitive in $\mathcal{G}^{(1)}$, we find the color difference to its NN $g_{j^*}^{(2)}$ in $\mathcal{G}^{(2)}$ and take the median over all primitives, weighted by the geometric kernel response:
\begin{equation}
\sigma_c^2 = \mathrm{median}\left(\left\{w_i \cdot \left\| \mathbf{c}_i - \mathbf{c}_{j^*} \right\|^2 \right\}_{i\in|\mathcal{G}|}\right) 
\;\;;\; w_i = k_{\text{geo}}(i, j^*).
\label{eq:sigmac}
\end{equation}
computed bidirectionally ($\mathcal{G}^{(1)} \to \mathcal{G}^{(2)}$ and $\mathcal{G}^{(2)} \to \mathcal{G}^{(1)}$) and averaged for symmetry. Weighting by $k_{\text{geo}}$ ensures only geometrically well-matched pairs contribute, suppressing contamination from genuinely changed regions. The median absorbs the global capture-side offset (which shifts the bulk of the distribution) and remains insensitive to the tail of genuinely changed pairs. We then adapt the bandwidth per primitive by spatial footprint:
\begin{equation}
\sigma_{c,i}^2 = \sigma_c^2 \cdot \max\!\left( h_i^2 / \tilde{h}^2,\; 1 \right),
\quad h_i^2 = \mathrm{tr}\!\left(\tilde{\bm{\Sigma}}_i^{\text{eff}}\right),
\quad \tilde{h}^2 = \mathrm{median}\left(\left\{h_k^2\right\}_{k\in|\mathcal{G}|}\right).
\label{eq:adaptive_sigma}
\end{equation}

Larger primitives aggregate appearance over a broader area and warrant proportionally wider bandwidth; floor at $\sigma_c^2$ prevents small, well-localized primitives from receiving an overly tight bandwidth.

\paragraph{Appearance kernel.}
Using our color bandwidth to capture photometric drift, we score color similarity with an isotropic RBF over the DC color $\mathbf{c}_i$:
\begin{equation}
k_{\text{app}}(g_i, g_j) = \exp\!\left( -\frac{\left\| \mathbf{c}_i - \mathbf{c}_j \right\|^2}{2 \, \sigma_{c,i}^2} \right).
\label{eq:kapp}
\end{equation}
Restricting comparison to the DC term discards higher-order SH coefficients that encode view-dependent effects (\S\ref{sec:prelim}), which would otherwise register as spurious appearance change; this matches the empirical finding of MV3DCD~\cite{galappaththige2025multi}. The per-primitive appearance change score is:
\begin{equation}
\delta_i^{\text{app}} = 1 - \max_{j \in \mathcal{N}(i)} k_{\text{app}}(g_i, g_j).
\label{eq:dapp}
\end{equation}
A high $\delta_i^{\text{app}}$ indicates that no primitive in the neighborhood $\mathcal{N}(i)$ explains $g_i$'s color once photometric drift has been absorbed, which we interpret as evidence of appearance change.

\subsubsection{Aggregating Per-Primitive Scores and Rendering}
\label{sec:agg}
After applying the geometric and appearance kernels, each primitive carries two change scores,  $(\delta_i^{\text{geo}},\delta_i^{\text{app}})$, computed bidirectionally for per-scene change information. We weight each score by a confidence term, render 2D maps at inference viewpoints, and combine the scenes into a change map.

\paragraph{Confidence weighting.} 
As foreshadowed in \S\ref{sec:geo}, observation uncertainty enters our pipeline at two stages. The geometric kernel absorbs observation uncertainty into the \emph{similarity computation}: FIM-based covariance inflation widens the matching tolerance for under-observed primitives so their position uncertainty does not register as change. The render step absorbs observation uncertainty into the \emph{aggregation}: a per-primitive confidence weight $\omega_i$ scales each primitive's contribution to the 2D map, so under-observed regions inform the final output proportionally to how well they were captured by the cameras.
Without the first, drift on uncertain primitives produces false positives; without the second, all primitives contribute equally regardless of how well-determined they are.
We analyze the two stages in \S\ref{sec:ablations} and show that they compose without redundancy. We define $\omega_i$ from the same FIM (Eq.~\ref{eq:fim}), centered on each scene's own reference:
\begin{equation}
\omega_i = \sigma\!\left( \log \mathrm{tr}(\mathbf{H}_i) - \log Q_{0.25}\!\left(\{\mathrm{tr}(\mathbf{H}_k)\}_{k \in |\mathcal{G}}|\right) \right),
\label{eq:wi}
\end{equation}
where $\sigma(\cdot)$ is the sigmoid and $\mathrm{tr}(\mathbf{H}_i)$ summarizes observability across the three spatial axes. The 25th-percentile reference suppresses the bottom quartile of each scene's primitives below $\omega_i = 0.5$, with the remainder passing through above. Because $\omega_i$ depends on each primitive's relative ranking within its own scene rather than absolute information level, it is independent of the COLMAP~\cite{schonberger2016structure} reconstruction scale, and a single formula applies across reconstructions of varying scale and density.

\paragraph{Rendering change maps.}
For each scene $s \in \{1, 2\}$ we form a saturated, observability-weighted per-primitive change score, scored bi-directionally, 
\begin{equation}
\delta_i^{(s)} = \omega_i \cdot \min\!\left( \delta_i^{\text{geo},(s)} + \delta_i^{\text{app},(s)} ,\; 1 \right),
\label{eq:dsat}
\end{equation}
and render it to a 2D map $\mathcal{M}^{(s)}$ at the inference viewpoint via alpha-compositing~\cite{kerbl20233d}, treating $\delta_i^{(s)}$ as a per-primitive scalar. The final change map is the pixel-wise maximum $\mathcal{M} = \max(\mathcal{M}^{(1)}, \mathcal{M}^{(2)})$.

\paragraph{Disambiguating structural from surface-only change.}
Our kernels provide per-primitive $(\delta_i^{\text{geo}}, \delta_i^{\text{app}})$, which we exploit to disambiguate structural from surface-only change. By construction, the geometric kernel detects only structural change: it fires on geometric mismatches. The appearance kernel, however, is ambiguous: it fires on any color mismatch within $\mathcal{N}(i)$, whether from surface-only recoloring or from a structural change whose revealed background differs from the previous foreground. The two cases are distinguishable because the second activates both kernels simultaneously, while the first activates only the appearance kernel. The residual $\max(\delta_i^{\text{app}} - \delta_i^{\text{geo}}, 0)$ therefore indicates the surface-only component. Rendering $\delta_i^{\text{geo}}$ and this residual independently yields two maps $\mathcal{M}^{\text{struct}}, \mathcal{M}^{\text{surf}}$, and for each pixel detected as changed in $\mathcal{M}$, we assign its type by
\begin{equation}
\text{label} = \arg\max\!\left( \mathcal{M}^{\text{struct}},\; \mathcal{M}^{\text{surf}} \right).
\label{eq:disambig}
\end{equation}
This change classification requires no manual annotations and no additional auxiliary model, and emerges directly from holding the two kernels separate.
\section{Experimental Analysis}
\label{sec:exp}

\textbf{Dataset.} We evaluate on PASLCD~\cite{galappaththige2025multi} following prior multi-view SCD literature~\cite{galappaththige2025changes, galappaththige2025multi}. PASLCD is a real-world benchmark of 10 scenes (indoor and outdoor) captured under both similar and different lighting, yielding 20 instances. Scenes contain multiple concurrent surface- and object-level changes alongside distractors such as shadows, reflections, and illumination shifts, captured from independently traversed trajectories. As an additional contribution, we have hand-annotated each ground-truth change pixel as \emph{structural} (geometric mismatch) or \emph{surface-level} (appearance-only), enabling evaluation of the change classification of \S\ref{sec:agg}. Of all changed pixels, $84\%$ are structural and $16\%$ are surface-level. We will release these annotations upon acceptance.

\textbf{Baselines.} We compare against the strongest representatives from each prior paradigm: supervised pairwise~\cite{cyws2d}, label-free pairwise~\cite{kim2025towards}, pose-agnostic anomaly detection~\cite{liu2024splatpose+}, and multi-view~\cite{galappaththige2025multi, lu20253dgs, galappaththige2025changes, zhou2026d}. Pairwise methods receive rendered aligned viewpoints, which simplifies their task.

\textbf{Metrics.} Following SCD literature~\cite{galappaththige2025changes, galappaththige2025multi, alcantarilla2018street, lu20253dgs, lin2025robust, cyws3d}, we report mIoU and F1 on changed pixels. Change maps $\mathcal{M} \in [0,1]$ are binarized at a fixed midpoint threshold of $0.5$ with no per-scene tuning advantage; we additionally report an \emph{Oracle} setting with per-scene optimal thresholds as the scoring upper bound. For change classification (\S\ref{sec:disambig}), we use balanced accuracy~\cite{brodersen2010balanced} as the primary metric and per-class precision/recall as secondary metrics, accounting for the structural-surface class imbalance.

\subsection{Main Results}
\label{sec:main_results}
 
\begin{table}[t]
\caption{Quantitative results on PASLCD~\cite{galappaththige2025multi}. \GSDiff{} is the only method that reaches state-of-the-art performance \emph{without external learned features} and with multi-view (MV) consistency as an \emph{inherent property} rather than a learned objective. Best in \textbf{bold}, second best \underline{underlined}.}
\label{tab:main}
\vspace{-3pt}
\centering
\small
\setlength{\tabcolsep}{5pt}
\renewcommand{\arraystretch}{0.85}
\begin{tabular}{l cc cc}
\toprule
Method & No Learned Feat. & MV Consistency & mIoU $\uparrow$ & F1 $\uparrow$ \\
\midrule
CYWS-2D~\citep{cyws2d}                  & \XSolidBrush  & --              & 0.273 & 0.398 \\
GeSCD~\citep{kim2025towards}            & \XSolidBrush  & --              & 0.477 & 0.611 \\
SplatPose+~\citep{liu2024splatpose+}    & \XSolidBrush  & --              & 0.237 & 0.358 \\
SCAR-3D~\citep{zhou2026d}               & \XSolidBrush  & Learned         & 0.191 & 0.289 \\
3DGS-CD~\citep{lu20253dgs}              & \XSolidBrush  & Learned         & 0.209 & 0.339 \\
MV3DCD~\citep{galappaththige2025multi}  & \XSolidBrush  & Learned         & 0.478 & 0.628 \\
O-SCD~\citep{galappaththige2025changes} & \XSolidBrush  & Learned         & \underline{0.552} & \underline{0.694} \\
\midrule
\textbf{\GSDiff{}}                      & \CheckmarkBold & by Construction & \textbf{0.644} & \textbf{0.758} \\
\GSDiff{} (Oracle)                      & \CheckmarkBold & by Construction & 0.669 & 0.779 \\
\bottomrule
\end{tabular}
\end{table}

\begin{figure}[t]
  \centering
  \vspace{-3pt}
  \includegraphics[width=\linewidth]{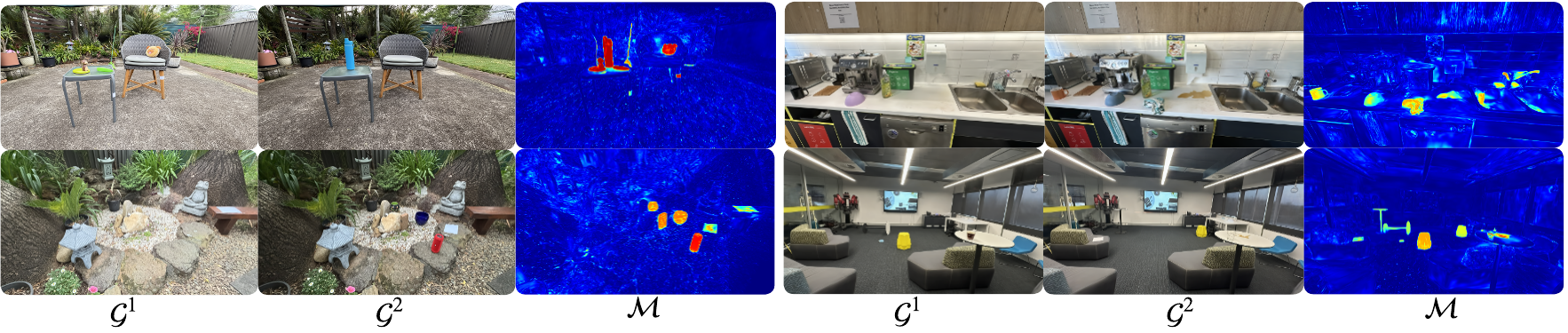}
  \vspace{-20pt}
  \caption{Qualitative results on PASLCD~\cite{galappaththige2025multi}. For each scene: rendered views from reference $\mathcal{G}^{(1)}$ (left), inference $\mathcal{G}^{(2)}$ (center), and change map $\mathcal{M}$ (right). \GSDiff{} produces sharply localized responses to both structural and surface-level changes across indoor and outdoor scenes.}
  \label{fig:vis}
  \vspace{-6pt}
\end{figure}

Table~\ref{tab:main} reports performance averaged over all 20 instances in PASLCD. \GSDiff{} establishes a new state of the art, improving mIoU by $\sim\textbf{17\%}$ over the strongest prior method while using only the native attributes of Gaussian primitives: position, covariance, and color. Every baseline relies on features of foundation models~\cite{oquab2023dinov2,ravi2024sam} and learned aggregation stage~\cite{galappaththige2025changes, galappaththige2025multi, zhou2026d} (except for pairwise methods); we exceed all of them through principled handling of reconstruction drift alone, providing direct empirical evidence for our first claim on the representational sufficiency of Gaussian primitives for multi-view SCD. The narrow Fixed-Oracle gap ($0.025$ mIoU) shows that our continuous scores are well-calibrated and do not require per-scene threshold tuning. Figure~\ref{fig:vis} provides a representative set of qualitative results, and additional visualizations are present in Appendix~\ref{app:vis}.

\subsection{Component Analysis}
\label{sec:ablations}

\begin{table}[t]
\caption{Component analysis on PASLCD~\cite{galappaththige2025multi}. Each row introduces one design choice cumulatively. Phases: \emph{Na\"ive} (raw nearest-neighbor comparison), \emph{Kernel} (geometric and appearance kernels with fixed parameters), \emph{Drift} (representation ambiguity and observation uncertainty across kernels).}
\label{tab:ablation}
\centering
\footnotesize
\setlength{\tabcolsep}{6pt}
\begin{tabular}{@{}l l c c c@{}}
\toprule
Phase & Variant & mIoU & Oracle mIoU & Relative Gap (\%) \\
\midrule
\emph{Na\"ive} & Euclidean NN (position + color) & 0.110 & 0.285 & 61.4 \\
 & Normalized Mahalanobis, raw $\bm{\Sigma}$ + Euclidean NN color & 0.034 & 0.266 & 87.2 \\
\midrule
\emph{Kernel} & Unnormalized Mahalanobis, raw $\bm{\Sigma}$ + Euclidean NN color & 0.096 & 0.415 & 76.9 \\
 & Unnormalized Mahalanobis + fixed RBF color ($\sigma_c{=}0.5$) & 0.102 & 0.422 & 75.8 \\
\midrule
\emph{Drift} & + representation ambiguity inflation $\mathbf{U}_i$ (\S\ref{sec:geo}) & 0.258 & 0.536 & 51.9 \\
 & + observation uncertainty inflation via FIM (\S\ref{sec:geo}) & 0.465 & 0.611 & 23.9 \\
 & + data-driven appearance bandwidth $\sigma_c$ (\S\ref{sec:app}) & 0.537 & 0.629 & 14.6 \\
 & + confidence weighting $\omega_i$ (\S\ref{sec:agg}) & \textbf{0.644} & \textbf{0.669} & \textbf{3.7} \\
\bottomrule
\end{tabular}
\end{table}

Table~\ref{tab:ablation} analyzes each component of \GSDiff{} cumulatively. Three observations stand out. \textbf{(1) Normalized Mahalanobis is numerically degenerate}: the standard form (row 2) under performs Euclidean NN (row 1) as its determinant prefactor penalizes co-located primitives whose covariance magnitudes differ across reconstructions (\S\ref{sec:geo}); dropping it (row 3) recovers $\sim\!3\times$ the mIoU. \textbf{(2) Drift modeling is the dominant component towards strong performance}: the drift-modeling phase lifts Fixed mIoU by $+0.542$, with geometric components contributing $+0.363$ jointly and appearance bandwidth plus observability weighting contributing the remaining $+0.179$; this validates our second claim that reconstruction drift, not feature representation, is the obstacle to direct primitive comparison. \textbf{(3) Observability is primarily a calibration mechanism}: FIM injection's Fixed-threshold gain is nearly $3\times$ its Oracle gain (row 5 to 6) -- separability was largely present after representation-ambiguity inflation, but the score distribution was poorly calibrated; the remaining drift components close the Fixed--Oracle gap to within $95\%$ of the Oracle ceiling. An extended analysis of data-driven quantile-sensitive is presented in Appendix~\ref{app:quantile-sensitivity}.

\subsection{Distinguishing Structural from Surface-Level Changes}
\label{sec:disambig}

Table~\ref{tab:disambig} shows the \GSDiff{} change classification approach achieves a balanced accuracy of $0.87$ under both Fixed and Oracle thresholds. Per-class recall is high in both classes ($0.96$ structural, $0.77$ surface-level), confirming the kernel-residual decomposition genuinely captures surface-level change. The asymmetric per-class precisions ($0.97$ vs $0.73$) reflect the underlying $84\%/16\%$ class imbalance: even small misrouting from the dominant structural class floods the smaller surface-level prediction set, suppressing surface-level precision regardless of routing accuracy. The near-identical Fixed and Oracle results further indicate the routing is decoupled from detection threshold tuning, validating the disentangling claim of \S\ref{sec:agg} -- holding the geometric and appearance kernels separate operating directly in primitive space exposes change \emph{type} as an emergent property, without supervision or external models. We show qualitative examples of this change classification in Appendix C.

\begin{table}[t]
\caption{Disambiguation routing on PASLCD~\cite{galappaththige2025multi}. Balanced accuracy~\cite{brodersen2010balanced} is the primary metric; structural/surface-level precision and recall are reported per class. Existing baselines produce a single change score per pixel and cannot natively disambiguate change against structural vs surface-only; \GSDiff{}'s primitive-space comparison enables this and requires no annotations or auxiliary models.}
\label{tab:disambig}
\centering
\footnotesize
\setlength{\tabcolsep}{6pt}
\renewcommand{\arraystretch}{0.9}
\begin{tabular}{l c cc cc}
\toprule
Thresholding & Balanced Accuracy & \multicolumn{2}{c}{Structural} & \multicolumn{2}{c}{Surface-level} \\
\cmidrule(lr){3-4} \cmidrule(lr){5-6}
& & Precision & Recall & Precision & Recall \\
\midrule
Fixed ($0.5$) & 0.868 & 0.970 & 0.961 & 0.725 & 0.774 \\
Oracle        & 0.866 & 0.968 & 0.959 & 0.726 & 0.773 \\
\bottomrule
\end{tabular}
\end{table}

\section{Limitations}
\label{sec:limitations}

\GSDiff{} treats reconstruction as an upstream black box: scoring quality is bounded by reconstruction quality, and pose errors from COLMAP propagate into both the kernel metric and the observability term. Geometrically accurate 3DGS is an active research area~\cite{chen2024pgsr, guedon2024sugar, Huang2DGS2024}, and \GSDiff{} benefits directly from advances there: cleaner geometric reconstructions sharpen the geometric kernel, sharpening the disambiguation routing. These limitations are inherited from the standard setup~\cite{galappaththige2025multi, lu20253dgs} and are orthogonal to our primitive-space contribution, but they bound the regime in which the method is reliable.
A second limitation lies in the photometric bandwidth $\sigma_c$, which absorbs color variation from both representation non-uniqueness and capture-condition offsets. This trade-off is favorable in inspection settings where capture-side changes are not of interest, but it could also absorb subtle surface-level changes. Inverse-rendering 3DGS variants~\cite{gao2024relightable, chen2025gigs} that separate environment lighting and material albedo offer a natural remedy: comparing on albedo alone removes the capture-side component from $\sigma_c$ and tightens the appearance kernel. We see this as a clean integration point for future work and orthogonal to the primitive-space contribution itself.

\section{Conclusion}
\label{sec:conclusion}
\GSDiff{} performs multi-view SCD directly in primitive space, achieving state-of-the-art performance while unlocking the ability to reason about \emph{what kind of} change occurred -- separating structural from surface-level appearance changes without supervision or auxiliary models. This combination of accuracy and interpretability is especially valuable for long-term monitoring applications such as heritage preservation~\cite{dahaghin2024gaussian} and industrial asset management~\cite{galappaththige2025multi}, where understanding why a region was flagged matters as much as flagging it. We see this work as the first step toward a new paradigm in SCD, and we hope it encourages the community to explore how far primitive-based change detection can push the reliability and capability of multi-view SCD.

\section*{Acknowledgment}
This work was supported by the Australian Research Council Research Hub in Intelligent Robotic Systems for Real-Time Asset Management (ARIAM) (IH210100030) and Abyss Solutions. C.J., N.S., and D.M. also acknowledge ongoing support from the QUT Centre for Robotics.


{\small
\bibliographystyle{ieeenat_fullname}
\bibliography{main_v2}
}

\newpage
\appendix

\section{\GSDiff{} Algorithm}
\label{app:algorithm}

We provide pseudo-code for the full \GSDiff{} pipeline (Algorithm~\ref{alg:gsdiff}). Reconstruction is treated as upstream input: we assume two independently built 3DGS reconstructions $\mathcal{G}^{(1)}, \mathcal{G}^{(2)}$ with COLMAP-estimated camera poses $\mathcal{C}^{(1)}, \mathcal{C}^{(2)}$ registered to a common frame, frustum-filtered to the shared observable region (\S\ref{sec:method-core}). \GSDiff{} then proceeds in four stages: geometric drift modeling (\S\ref{sec:geo}), kernel scoring on geometry and appearance (\S\ref{sec:geo}, \S\ref{sec:app}), aggregation and rendering (\S\ref{sec:agg}), and disambiguation (\S\ref{sec:agg}). All quantities derived from inter-reconstruction statistics -- the representation-ambiguity scales $u_n, u_t$, the appearance bandwidth $\sigma_c$ -- are computed bidirectionally ($\mathcal{G}^{(1)}{\to}\mathcal{G}^{(2)}$ and $\mathcal{G}^{(2)}{\to}\mathcal{G}^{(1)}$) and averaged for symmetry; per-primitive quantities ($\tilde{\bm{\Sigma}}_i^{\text{eff}}$, $\omega_i$, $\delta_i^{\text{geo}}$, $\delta_i^{\text{app}}$) are computed once per primitive in each reconstruction.

\begin{algorithm}[H]
\caption{\GSDiff{}: Primitive-Space Scene Change Detection}
\label{alg:gsdiff}
\begin{algorithmic}[1]
\Require Reconstructions $\mathcal{G}^{(1)}, \mathcal{G}^{(2)}$ with poses $\mathcal{C}^{(1)}, \mathcal{C}^{(2)}$ in a common frame, frustum-filtered to the shared observable region
\Ensure Binary change mask $\mathcal{M}$; per-pixel structural/surface labels

\Statex \textbf{Stage 1: Geometric drift modeling (\S\ref{sec:geo})}
\For{$s \in \{1, 2\}$, primitive $i \in \mathcal{G}^{(s)}$}
    \State $j^* \gets$ Euclidean nearest neighbor of $i$ in $\mathcal{G}^{(\bar{s})}$ 
    \State Compute $d_n^{(i)}, d_t^{(i)}$ via Eq.~\ref{eq:repamb}
    \State Compute $\mathbf{H}_i$ over visible cameras $\mathcal{C}_i^{\text{vis}}$ \Comment{Eq.~\ref{eq:fim}}
\EndFor
\State $u_n^2, u_t^2 \gets$ symmetric average of $Q_{0.75}$ over $\mathcal{G}^{(1)}{\to}\mathcal{G}^{(2)}$ and $\mathcal{G}^{(2)}{\to}\mathcal{G}^{(1)}$
\For{$s \in \{1, 2\}$, primitive $i \in \mathcal{G}^{(s)}$}
    \State $\tilde{\bm{\Sigma}}_i \gets \bm{\Sigma}_i + u_t^2 \mathbf{I}_3 + (u_n^2 - u_t^2)\mathbf{n}_i\mathbf{n}_i^\top$ \Comment{Eq.~\ref{eq:repamb}}
    \State $\alpha^{(s)} \gets$ FIM-injection scale on $\mathcal{G}^{(s)}$ \Comment{numerator of Eq.~\ref{eq:obsunc}}
    \State $\tilde{\bm{\Sigma}}_i^{\text{eff}} \gets \tilde{\bm{\Sigma}}_i + \alpha^{(s)} \mathbf{H}_i^{+}$ \Comment{Eq.~\ref{eq:obsunc}}
\EndFor

\Statex \textbf{Stage 2: Kernel scoring (\S\ref{sec:geo}, \S\ref{sec:app})}
\State $\sigma_c^2 \gets$ symmetric weighted-median appearance bandwidth \Comment{Eq.~\ref{eq:sigmac}}
\For{$s \in \{1, 2\}$, primitive $i \in \mathcal{G}^{(s)}$}
    \State $\mathcal{N}(i) \gets \{j \in \mathcal{G}^{(\bar{s})} : \|\bm{\mu}_i - \bm{\mu}_j\| \leq 3\sqrt{\lambda_{\max}(\tilde{\bm{\Sigma}}_i^{\text{eff}})}\}$ \Comment{Eq.~\ref{eq:neighbors}}
    \State $\delta_i^{\text{geo}} \gets 1 - \max_{j \in \mathcal{N}(i)} k_{\text{geo}}(g_i, g_j)$ \Comment{Eq.~\ref{eq:dgeo}}
    \State $\sigma_{c,i}^2 \gets \sigma_c^2 \cdot \max(h_i^2/\tilde{h}^2, 1)$ \Comment{Eq.~\ref{eq:adaptive_sigma}}
    \State $\delta_i^{\text{app}} \gets 1 - \max_{j \in \mathcal{N}(i)} k_{\text{app}}(g_i, g_j)$ \Comment{Eq.~\ref{eq:dapp}}
\EndFor

\Statex \textbf{Stage 3: Aggregation and rendering (\S\ref{sec:agg})}
\For{$s \in \{1, 2\}$, primitive $i \in \mathcal{G}^{(s)}$}
    \State $\omega_i \gets \sigma\big(\log \mathrm{tr}(\mathbf{H}_i) - \log Q_{0.25}(\{\mathrm{tr}(\mathbf{H}_k)\}_{k \in \mathcal{G}^{(s)}})\big)$ \Comment{Eq.~\ref{eq:wi}}
    \State $\delta_i^{(s)} \gets \omega_i \cdot \min(\delta_i^{\text{geo}} + \delta_i^{\text{app}}, 1)$ \Comment{Eq.~\ref{eq:dsat}}
\EndFor
\State $\mathcal{M}^{(s)} \gets \textsc{AlphaCompositeRender}(\mathcal{G}^{(s)}, \delta_i^{(s)})$ for $s \in \{1, 2\}$
\State $\mathcal{M} \gets \max(\mathcal{M}^{(1)}, \mathcal{M}^{(2)})$, binarized at $0.5$ \Comment{final change map}

\Statex \textbf{Stage 4: Disambiguation (\S\ref{sec:agg})}
\State $\mathcal{M}^{\text{struct}} \gets \textsc{AlphaCompositeRender}(\mathcal{G}, \delta_i^{\text{geo}})$
\State $\mathcal{M}^{\text{surf}} \gets \textsc{AlphaCompositeRender}(\mathcal{G}, \max(\delta_i^{\text{app}} - \delta_i^{\text{geo}}, 0))$
\For{each pixel $p$ with $\mathcal{M}_p > 0.5$}
    \State $\text{label}(p) \gets \arg\max(\mathcal{M}^{\text{struct}}_p, \mathcal{M}^{\text{surf}}_p)$ \Comment{Eq.~\ref{eq:disambig}}
\EndFor
\end{algorithmic}
\end{algorithm}

\paragraph{Implementation notes.}
Nearest-neighbor and ball queries use spatial KD-trees, giving $O(N \log N)$ per scene where $N$ is the primitive count. The FIM accumulation in Eq.~\ref{eq:fim} is per-primitive over its frustum-visible cameras and is batched across primitives. The two final renderings ($\mathcal{M}^{(s)}$ for the change map; $\mathcal{M}^{\text{struct}}, \mathcal{M}^{\text{surf}}$ for disambiguation) reuse the standard 3DGS rasterizer~\cite{kerbl20233d} with $\delta_i^{(s)}$, $\delta_i^{\text{geo}}$, and $\max(\delta_i^{\text{app}} - \delta_i^{\text{geo}}, 0)$ substituted for the per-primitive color in alpha-compositing. We use PGSR~\cite{chen2024pgsr} to build geometrically accurate reconstructions $\mathcal{G}^{(1)},\mathcal{G}^{(2)}$. All experiments are conducted on a single NVIDIA RTX4090 24GB GPU. 

\section{Analysis on Data-Driven Quantiles}
\label{app:quantile-sensitivity}

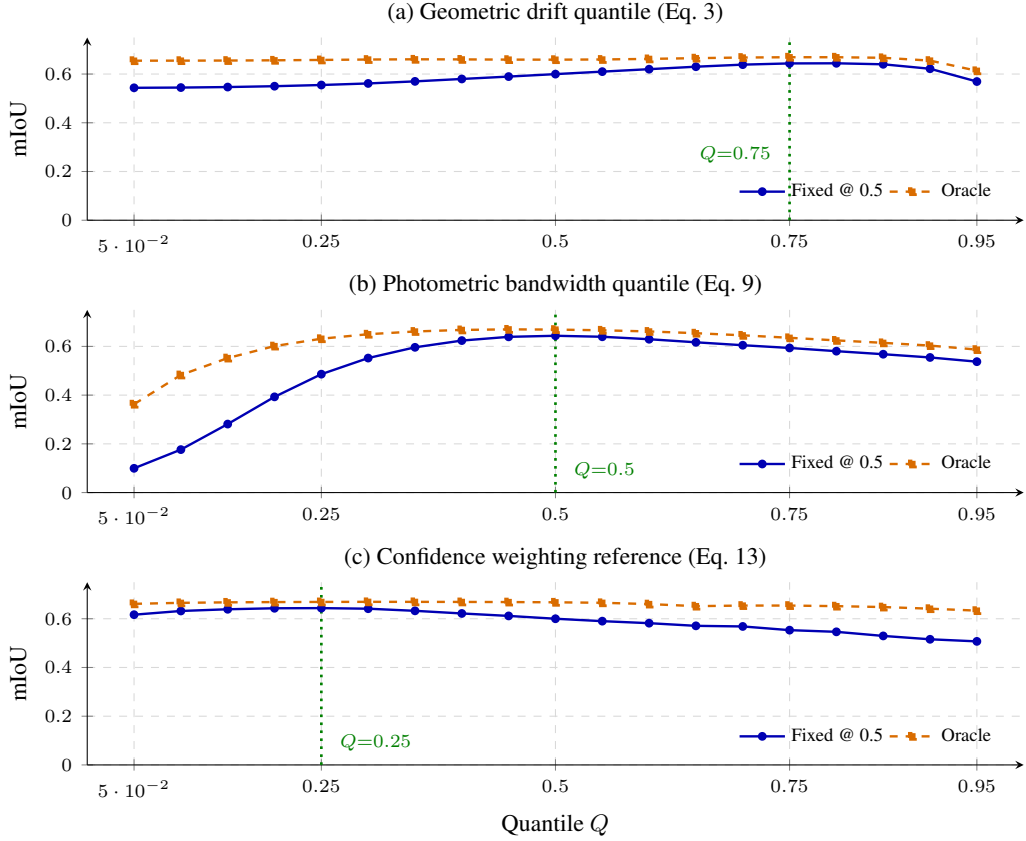
\begin{figure}[t]
\centering
\pgfplotsset{
  sweep axis/.style={
    width=\linewidth, height=4.0cm,
    xmin=0, xmax=1, ymin=0, ymax=0.75,
    xtick={0.05, 0.25, 0.5, 0.75, 0.95},
    xticklabel style={font=\scriptsize},
    yticklabel style={font=\scriptsize},
    xlabel style={font=\small},
    ylabel style={font=\small},
    title style={font=\small, yshift=-1ex},
    legend style={font=\scriptsize, draw=none, fill=none,
                  legend cell align=left, at={(0.98,0.05)}, anchor=south east, legend columns=2},
    grid=major, grid style={dashed, gray!30},
    axis lines=left,
    fixed style/.style={mark=*, mark size=1.2pt, color=blue!70!black, line width=0.9pt},
    oracle style/.style={mark=square*, mark size=1.2pt, color=orange!85!black, line width=0.9pt, dashed},
  },
}
\tikzset{
  chosen style/.style={dotted, line width=1.0pt, color=green!50!black},
}
\pgfplotsset{
  fixed style/.style={mark=*, mark size=1.2pt, color=blue!70!black, line width=0.9pt},
  oracle style/.style={mark=square*, mark size=1.2pt, color=orange!85!black, line width=0.9pt, dashed},
  chosen style/.style={dotted, line width=1.0pt, color=green!50!black},
}

\begin{tikzpicture}
\begin{groupplot}[
  group style={
    group size=1 by 3,
    vertical sep=7.5ex,
  },
  sweep axis,
]

\nextgroupplot[
  ylabel=mIoU,
  title={(a) Geometric drift quantile  (Eq.~\ref{eq:repamb})},
  xlabel={},
]
\addplot[fixed style] coordinates {
  (0.05,0.5435) (0.10,0.5445) (0.15,0.5466) (0.20,0.5501) (0.25,0.5550)
  (0.30,0.5614) (0.35,0.5699) (0.40,0.5798) (0.45,0.5896) (0.50,0.5997)
  (0.55,0.6099) (0.60,0.6200) (0.65,0.6301) (0.70,0.6386) (0.75,0.6436)
  (0.80,0.6441) (0.85,0.6400) (0.90,0.6217) (0.95,0.5693)
};
\addlegendentry{Fixed @ 0.5}
\addplot[oracle style] coordinates {
  (0.05,0.6548) (0.10,0.6550) (0.15,0.6554) (0.20,0.6564) (0.25,0.6579)
  (0.30,0.6596) (0.35,0.6605) (0.40,0.6601) (0.45,0.6592) (0.50,0.6590)
  (0.55,0.6598) (0.60,0.6622) (0.65,0.6652) (0.70,0.6678) (0.75,0.6690)
  (0.80,0.6690) (0.85,0.6667) (0.90,0.6552) (0.95,0.6134)
};
\addlegendentry{Oracle}
\draw[chosen style] (axis cs:0.75,0) -- (axis cs:0.75,0.75);
\node[anchor=south east, font=\scriptsize, color=green!50!black] at (axis cs:0.74,0.2) {$Q{=}0.75$ };

\vspace{20pt}
\nextgroupplot[
  ylabel=mIoU,
  title={(b) Photometric bandwidth quantile (Eq.~\ref{eq:sigmac})},
  xlabel={},
]
\addplot[fixed style] coordinates {
  (0.05,0.0996) (0.10,0.1764) (0.15,0.2812) (0.20,0.3929) (0.25,0.4860)
  (0.30,0.5521) (0.35,0.5962) (0.40,0.6239) (0.45,0.6392) (0.50,0.6436)
  (0.55,0.6395) (0.60,0.6294) (0.65,0.6168) (0.70,0.6048) (0.75,0.5936)
  (0.80,0.5804) (0.85,0.5679) (0.90,0.5546) (0.95,0.5373)
};
\addlegendentry{Fixed @ 0.5}
\addplot[oracle style] coordinates {
  (0.05,0.3628) (0.10,0.4827) (0.15,0.5521) (0.20,0.6013) (0.25,0.6315)
  (0.30,0.6500) (0.35,0.6612) (0.40,0.6676) (0.45,0.6699) (0.50,0.6690)
  (0.55,0.6661) (0.60,0.6613) (0.65,0.6542) (0.70,0.6454) (0.75,0.6353)
  (0.80,0.6246) (0.85,0.6146) (0.90,0.6033) (0.95,0.5869)
};
\addlegendentry{Oracle}
\draw[chosen style] (axis cs:0.5,0) -- (axis cs:0.5,0.75);
\node[anchor=south west, font=\scriptsize, color=green!50!black] at (axis cs:0.51,0.02) {$Q{=}0.5$};

\nextgroupplot[
  ylabel=mIoU,
  title={(c) Confidence weighting reference (Eq.~\ref{eq:wi})},
  xlabel={Quantile $Q$},
]
\addplot[fixed style] coordinates {
  (0.05,0.6166) (0.10,0.6319) (0.15,0.6389) (0.20,0.6430) (0.25,0.6436)
  (0.30,0.6412) (0.35,0.6325) (0.40,0.6217) (0.45,0.6116) (0.50,0.6000)
  (0.55,0.5901) (0.60,0.5818) (0.65,0.5709) (0.70,0.5684) (0.75,0.5532)
  (0.80,0.5463) (0.85,0.5295) (0.90,0.5157) (0.95,0.5071)
};
\addlegendentry{Fixed @ 0.5}
\addplot[oracle style] coordinates {
  (0.05,0.6612) (0.10,0.6653) (0.15,0.6673) (0.20,0.6684) (0.25,0.6690)
  (0.30,0.6694) (0.35,0.6693) (0.40,0.6690) (0.45,0.6685) (0.50,0.6675)
  (0.55,0.6655) (0.60,0.6602) (0.65,0.6517) (0.70,0.6541) (0.75,0.6541)
  (0.80,0.6517) (0.85,0.6480) (0.90,0.6413) (0.95,0.6333)
};
\addlegendentry{Oracle}
\draw[chosen style] (axis cs:0.25,0) -- (axis cs:0.25,0.75);
\node[anchor=south west, font=\scriptsize, color=green!50!black] at (axis cs:0.26,0.02) {$Q{=}0.25$ };

\end{groupplot}
\end{tikzpicture}

\caption{Sensitivity of \GSDiff{} to data-driven quantile choices: (a) representation-ambiguity drift scales, (b) appearance bandwidth, (c) observability-weighting reference. We sweep each quantile from $0.05$ to $0.95$ in $0.05$ steps on PASLCD; the green dotted line marks our \emph{a priori} principled choice (upper quartile, median, lower quartile, respectively). All three choices (green dotted line) sit within $0.01$ mIoU of the empirical Fixed-threshold optimum on PASLCD; Oracle mIoU is largely insensitive to quantile value (except for photometric bandwidth at $Q<0.3$), suggesting quantile choice primarily affects calibration rather than underlying separability.}
\label{fig:quantile-sweep}
\end{figure}

\GSDiff{} computes three data-driven quantities through quantiles of empirical distributions: the geometric drift scales $u_n, u_t$ as the upper quartile $Q_{0.75}$ of nearest-neighbor displacements (Eq.~\ref{eq:repamb}); the photometric bandwidth $\sigma_c$ as the median ($Q_{0.50}$) of weighted color differences (Eq.~\ref{eq:sigmac}); and the confidence-weighting reference as the lower quartile $Q_{0.25}$ of FIM traces (Eq.~\ref{eq:wi}). These three quantiles -- upper, median, lower -- correspond to the form of Tukey's standard five-number descriptive-statistics summary of a distribution~\cite{tukey1977exploratory}: the upper quartile bounds the bulk of typical drift while excluding the tail of genuinely changed regions; the median is the canonical robust location estimate; the lower quartile isolates the bottom-quartile tail of poorly observed primitives. We chose these three values \emph{a priori} on these statistical grounds.

To ablate the sensitivity of performance to these values, we sweep each quantile from $0.05$ to $0.95$ in steps of $0.05$ on PASLCD, holding all other components fixed. Figure~\ref{fig:quantile-sweep} reports Fixed-threshold mIoU (primary, threshold $0.5$) and Oracle mIoU (per-scene optimal threshold) for each sweep. In all three cases, the Oracle mIoU remains generally stable across quantile values (with the exception of the photometric bandwidth quantiles below 0.3, where performance decrements occur). In contrast, the Fixed-threshold performance shows some sensitivity to quantile value, indicating that quantile selection primarily affects the calibration of change scores against a fixed threshold rather than the underlying separability. 

As shown in Figure~\ref{fig:quantile-sweep}, our quantile selections sit at or within $0.01$ mIoU of the empirical optimum within fairly stable plateaus: the geometric drift quantile plateaus between $Q_{0.70}$ and $Q_{0.85}$; the photometric bandwidth quantile plateaus between $Q_{0.45}$ and $Q_{0.55}$; finally the confidence weighting reference quantile plateaus between $Q_{0.15}$ and $Q_{0.30}$. This suggests that the performance of \GSDiff{} has some robustness to exact quantile value selection, provided that the general logic (described above) is followed.

\section{Additional Visualizations of Change Maps}
\label{app:vis}
\begin{figure}[t]
  \centering
  \includegraphics[width=\linewidth]{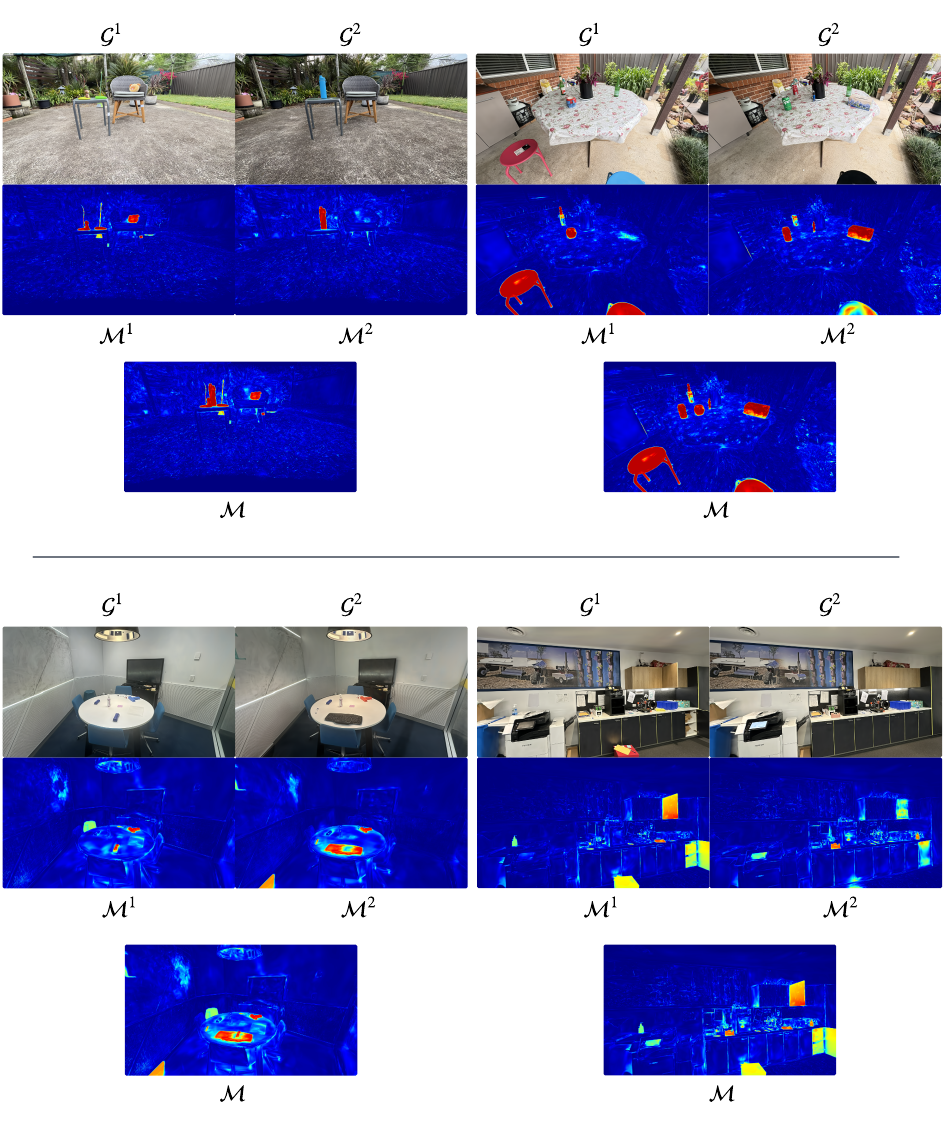}
  \vspace{-18pt}
  \caption{Qualitative results on PASLCD~\cite{galappaththige2025multi}. For each scene: rendered views from reference $\mathcal{G}^{(1)}$ (top-left), inference $\mathcal{G}^{(2)}$ (top-right), and change map for each individual scene $\mathcal{M}^1$ (bottom - left), $\mathcal{M}^2$ (bottom - right), and final change map $\mathcal{M}$ (bottom - center). \GSDiff{} produces sharply localized responses to both structural and surface-level changes across indoor and outdoor scenes. Our bidirectional scoring of change (\S\ref{sec:agg}) allows us to generate change maps for each reference and inference scenes individually. }
  \label{fig:vis_scene}
\end{figure}

\begin{figure}[t]
  \centering
  \includegraphics[width=\linewidth]{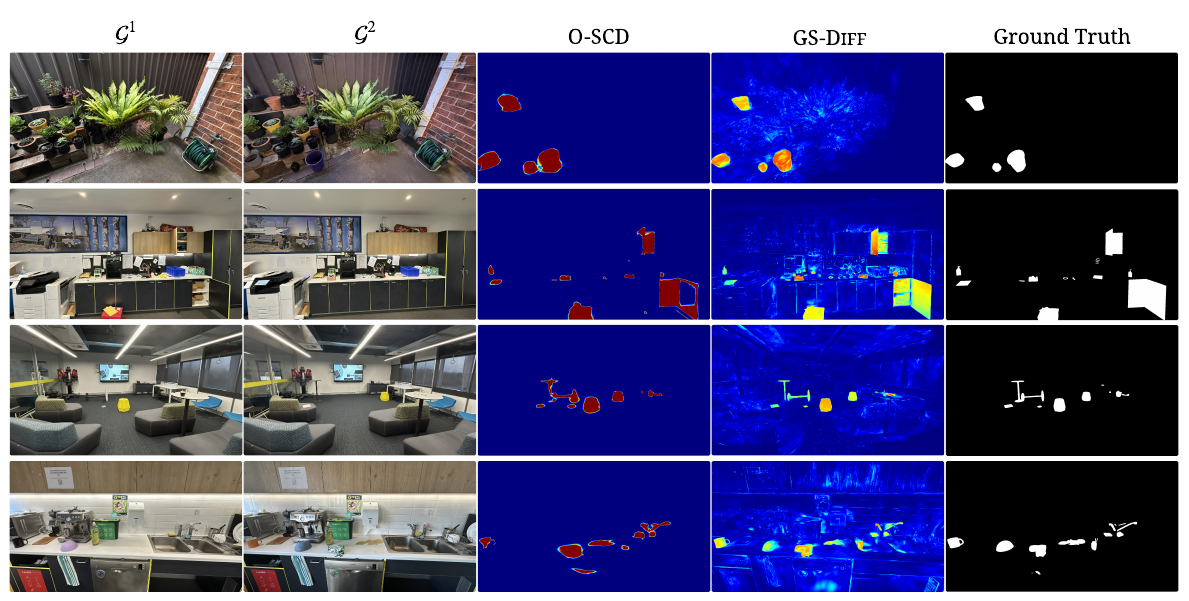}
  \vspace{-18pt}
  \caption{Qualitative comparison with O-SCD~\cite{galappaththige2025changes}, the strongest image-space baseline. \GSDiff{} produces more geometrically accurate change masks for two reasons. (1)~\emph{Patch granularity:} O-SCD and other image-space approaches compare features from an external foundation model~\cite{ravi2024sam} at its patch size ($14\times14$ or $16\times16$ in pixels), which sets a floor on spatial resolution; primitive-space comparison is limited only by the fidelity of the representation itself. (2)~\emph{Multi-view aggregation:} O-SCD's learned multi-view objective converges to a consensus across views, smoothing the boundary in any single view; \GSDiff{}'s per-primitive scores are multi-view consistent by construction and preserve view-specific sharpness. The bottom row additionally shows our kernels detecting a surface-level change between semantically similar objects (bowl recolored pink-to-blue), which O-SCD misses, as foundation-model features are largely invariant to such fine appearance changes in semantically similar objects~\cite{galappaththige2025changes,galappaththige2025multi}.  O-SCD's multi-view learning objective drives its outputs toward binary scores 0 or 1, producing the bimodal red/blue maps shown above; \GSDiff{}'s scores arise from kernel evaluations and remain continuous, exposing graded confidence and enabling the threshold-stable behavior reported in \S\ref{sec:main_results}.}

  \label{fig:oscd}
\end{figure}

\begin{figure}[t]
  \centering
  \includegraphics[width=\linewidth]{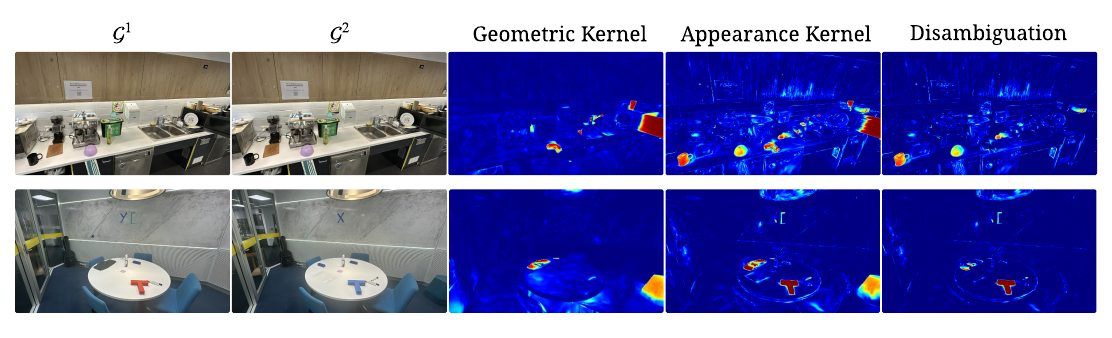}
  \vspace{-18pt}
  \caption{Qualitative results on kernel scores and disambiguation routing  (\S\ref{sec:agg}). Left to right: rendered views from reference $\mathcal{G}^1$ and inference $\mathcal{G}^2$ scenes: geometric kernel change score $\delta^{\text{geo}}$; appearance kernel change score $\delta^{\text{app}}$; and the residual $\max(\delta_i^{\text{app}} - \delta_i^{\text{geo}}, 0)$ used in structural versus sufrace-only disambiguation\S\ref{sec:agg}. Note that residual has a high score in surface only changes (Row 1: color change in mug, bowl, tea pack on the oven and coffee spill, Row 2: Color change in T shape object and drawing on the whiteboard), whereas geometric change scores are minimal in these regions.}

  \label{fig:disamb}
\end{figure}

We provide additional visualizations of our final change maps and the per-scene maps generated for each reference and inference scenes separately in Figure~\ref{fig:vis_scene}.Figure~\ref{fig:oscd} provides a qualitative comparison against the strongest image-space baseline, O-SCD~\cite{galappaththige2025changes}. Figure~\ref{fig:disamb} visualizes the per-kernel scores and the residual that drives the structural-versus-surface-only routing of \S\ref{sec:agg}.

\end{document}